\documentclass{article}
\usepackage{spconf,amsmath,graphicx}
\usepackage{times}
\usepackage{amssymb}
\usepackage{multirow}
\usepackage{booktabs}
\usepackage{xcolor}
\usepackage{url}
\usepackage[hidelinks]{hyperref}
\title{M2Retinexformer: Multi-Modal Retinexformer for Low-Light Image Enhancement}

\name{Youssef Aboelwafa, Hicham G.~Elmongui, Marwan Torki}
\address{Alexandria University, Egypt}

\begin{document}
\maketitle

\begin{abstract}

Low-light image enhancement is challenging due to complex degradations, including amplified noise, artifacts, and color distortion. While Retinex-based deep learning methods have achieved promising results, they primarily rely on single-modality RGB information. We propose M2Retinexformer (Multi-Modal Retinexformer), a novel framework that extends Retinexformer by incorporating depth cues, luminance priors, and semantic features within a progressive refinement pipeline. Depth provides geometric context that is invariant to lighting variations, while luminance and semantic features offer explicit guidance on brightness distribution and scene understanding. Modalities are extracted at multiple scales and fused through cross-attention, with adaptive gating dynamically balancing illumination-guided self-attention and cross-attention based on the reliability of auxiliary cues. Evaluations on the LOL, SID, SMID, and SDSD benchmarks demonstrate overall improvements over Retinexformer and recent state-of-the-art methods. Code and pretrained weights are available at \url{https://github.com/YoussefAboelwafa/M2Retinexformer}.

\end{abstract}

% \vspace{2mm}
\begin{keywords}
low-light image enhancement, Retinex theory, multi-modal learning, depth estimation, transformer
\end{keywords}

% \vspace{2mm}
\section{INTRODUCTION}
\label{sec:intro}

Low-light image enhancement is a challenging problem in image processing that aims to restore visibility and suppress corruptions in under-exposed images. Images captured under poor illumination conditions suffer from multiple degradations, including poor visibility, reduced contrast, amplified noise, and color distortion. These artifacts degrade perceptual quality and impair downstream vision tasks such as object detection, semantic segmentation, and recognition, all of which assume well-exposed inputs~\cite{exdark}.

\begin{figure}[!t]
	\raggedleft
	\begin{minipage}{\columnwidth}
        \centering
        \includegraphics[width=\linewidth]{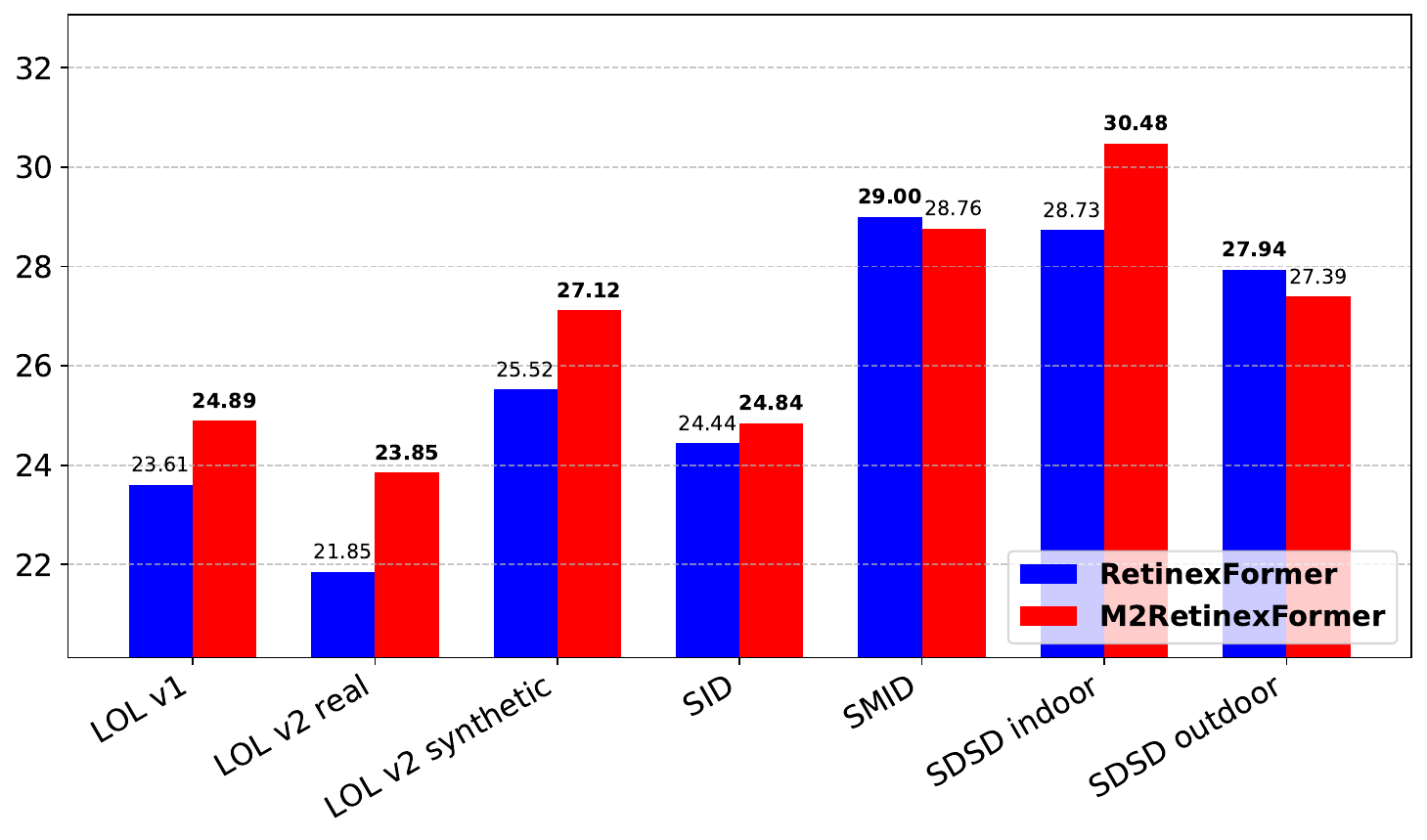}
    \end{minipage}
    \caption{\small The proposed M2Retinexformer achieves higher PSNR than the baseline Retinexformer on most evaluated datasets.
    }
    \label{fig:metrics_comparison}
\end{figure}

The Retinex theory~\cite{Land1971} provides a physical framework for addressing low-light enhancement by decomposing an image into reflectance and illumination components. Several deep learning methods have adopted this decomposition~\cite{retinex_net, kind, retinexformer, RetinexMamba}, with Retinexformer~\cite{retinexformer} achieving particularly strong results through its One-stage Retinex-based Framework and Illumination-Guided Transformer. 
% \\

\begin{figure}[t]
    \centering
    \begin{tabular}{c}
        \includegraphics[width=0.46\textwidth]{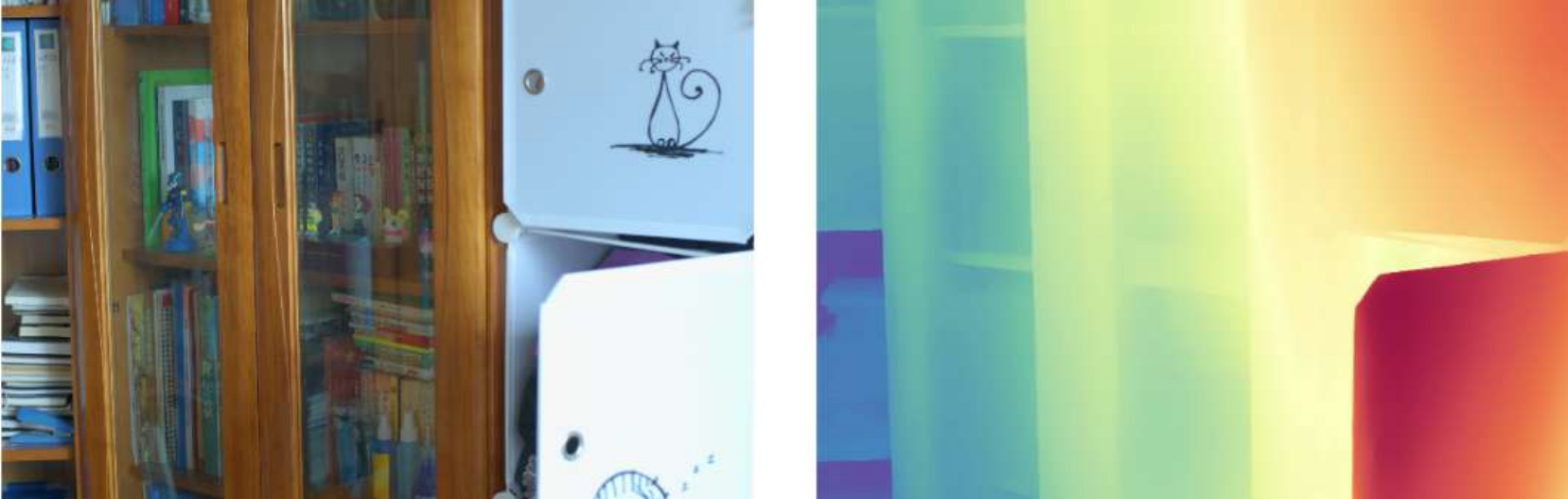} \\
        \small (a) Normal-light image depth \\[2mm]
        \includegraphics[width=0.46\textwidth]{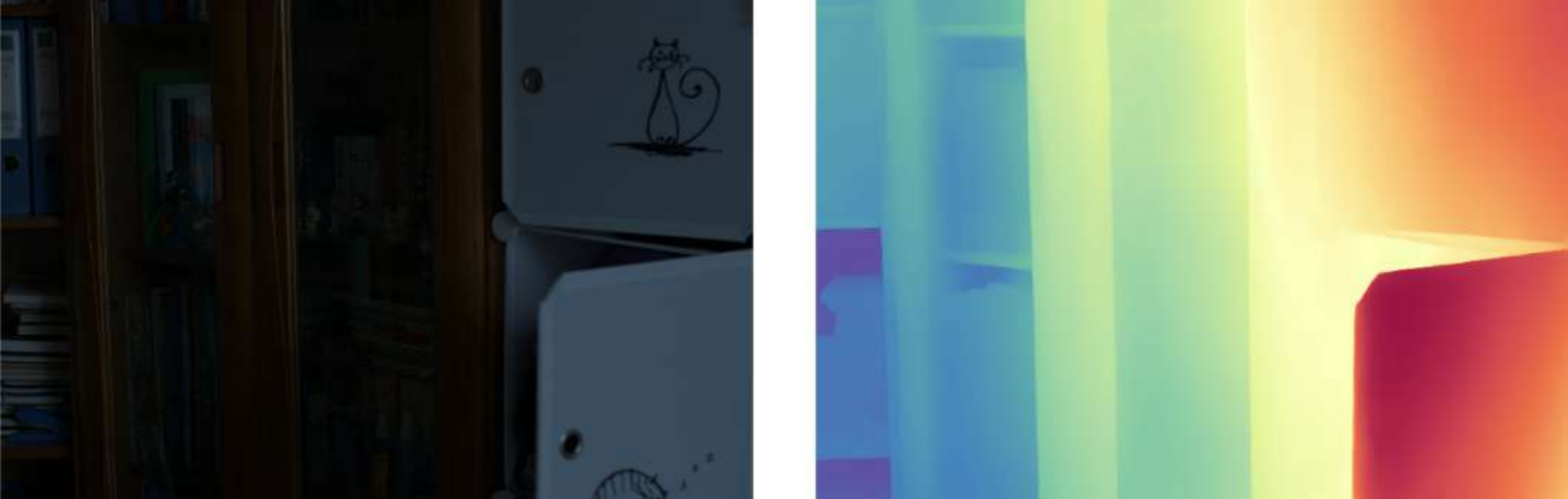} \\
        \small (b) Low-light image depth
    \end{tabular}
    \vspace{-2mm}
    \caption{\small The depth maps remain consistent under different illumination conditions, demonstrating that depth estimation is largely independent of image brightness.}
    \label{fig:depth_illumination}
\end{figure}

However, Retinexformer~\cite{retinexformer} relies exclusively on RGB information, which limits the network's ability to reason about scene geometry and the spatial distribution of light across surfaces. Motivated by this limitation, our work is based on three key observations:

\textbf{(i) Depth encodes geometric structure.}
As illustrated in Fig.~\ref{fig:depth_illumination}, depth maps remain largely consistent regardless of illumination. These geometric cues help distinguish between dark regions caused by distance, occlusion, or shadows. Depth helps disambiguate these cases by providing geometric information that is robust to brightness variations.

\textbf{(ii) Luminance and semantic features provide content-aware guidance.}
In Retinexformer, the illumination prior is extracted once at the beginning and concatenated with the RGB image, after which the network no longer needs this information.
In contrast, our approach maintains luminance features as a persistent modality and fuses them via cross-attention throughout the enhancement process. In addition, we propagate semantic features throughout the network to preserve natural colors, fine textures, and object boundaries.

\textbf{(iii) Cross-attention enables fusion of heterogeneous modalities.} Recent advances in multi-modal learning~\cite{modalformer} have demonstrated that cross-attention enables effective information exchange between heterogeneous modalities. 

\noindent Based on these observations, our contributions are summarized as follows:

\begin{itemize}
	\item We introduce \textbf{M2Retinexformer} that extends Retinexformer~\cite{retinexformer} by incorporating depth, luminance, and semantic features as auxiliary modalities through a Multi-Modal Cross-Attention Block (MMCAB) and an adaptive gating mechanism that balances self-attention and cross-attention based on auxiliary reliability. 
    The proposed design fuses heterogeneous modality features within a modular and extensible architecture, enabling flexible integration of additional modalities without modifying the core network.
    \vspace{-2mm}
    \item Through extensive analysis and ablation studies, we systematically investigate the contribution of each auxiliary modality and demonstrate their individual and combined effects on performance. 
    Experiments on LOL, SID, SMID, and SDSD benchmarks show that M2Retinexformer achieves improved performance over Retinexformer on the majority of evaluated datasets, as shown in Fig.~\ref{fig:metrics_comparison}.
\end{itemize}

\section{RELATED WORK}
\label{sec:related}
\vspace{-2mm}
\noindent\textbf{Classical Methods}: Retinex theory, introduced by Land~\cite{Land1971}, has shaped numerous enhancement algorithms. Classical approaches such as~\cite{ssr, msr, lime} rely on hand-crafted priors and assume that low-light images are corruption-free, leading to noise amplification and color distortion.

\noindent\textbf{Zero-Reference Methods}: Methods such as~\cite{guo2020zero, saeed2023lit} learn enhancement mappings directly from input images without paired supervision, typically using unpaired datasets.

\noindent\textbf{CNNs}: RetinexNet~\cite{retinex_net}, KinD~\cite{kind}, and URetinex-Net~\cite{uretinex} extend Retinex decomposition with CNNs. 

\noindent\textbf{Vision Transformers}: Restormer~\cite{restormer} and Uformer~\cite{uformer} introduced efficient self-attention mechanisms for image restoration. SNR-Net~\cite{snr_net} combines CNN and Transformer with signal-to-noise ratio guidance. Retinexformer~\cite{retinexformer} is the first single-stage transformer among Retinex-based methods, introducing Illumination-Guided Multi-head Self-Attention (IG-MSA). Retinexformer+~\cite{retinexformer+} extended this with multi-scale dilated convolutions and dual self-attention.

\noindent\textbf{State Space Model}: RetinexMamba~\cite{RetinexMamba} takes a different direction, replacing the transformer with a Mamba state-space model~\cite{gu2024mamba} to achieve linear complexity.

\noindent\textbf{Diffusion Models}: Recent diffusion-based methods such as~\cite{he2025reti, elkordi2026pwc} recast low-light enhancement as an iterative generative restoration process.

\begin{figure*}[htb]
	\centering
	\includegraphics[height=0.4\textheight, keepaspectratio]{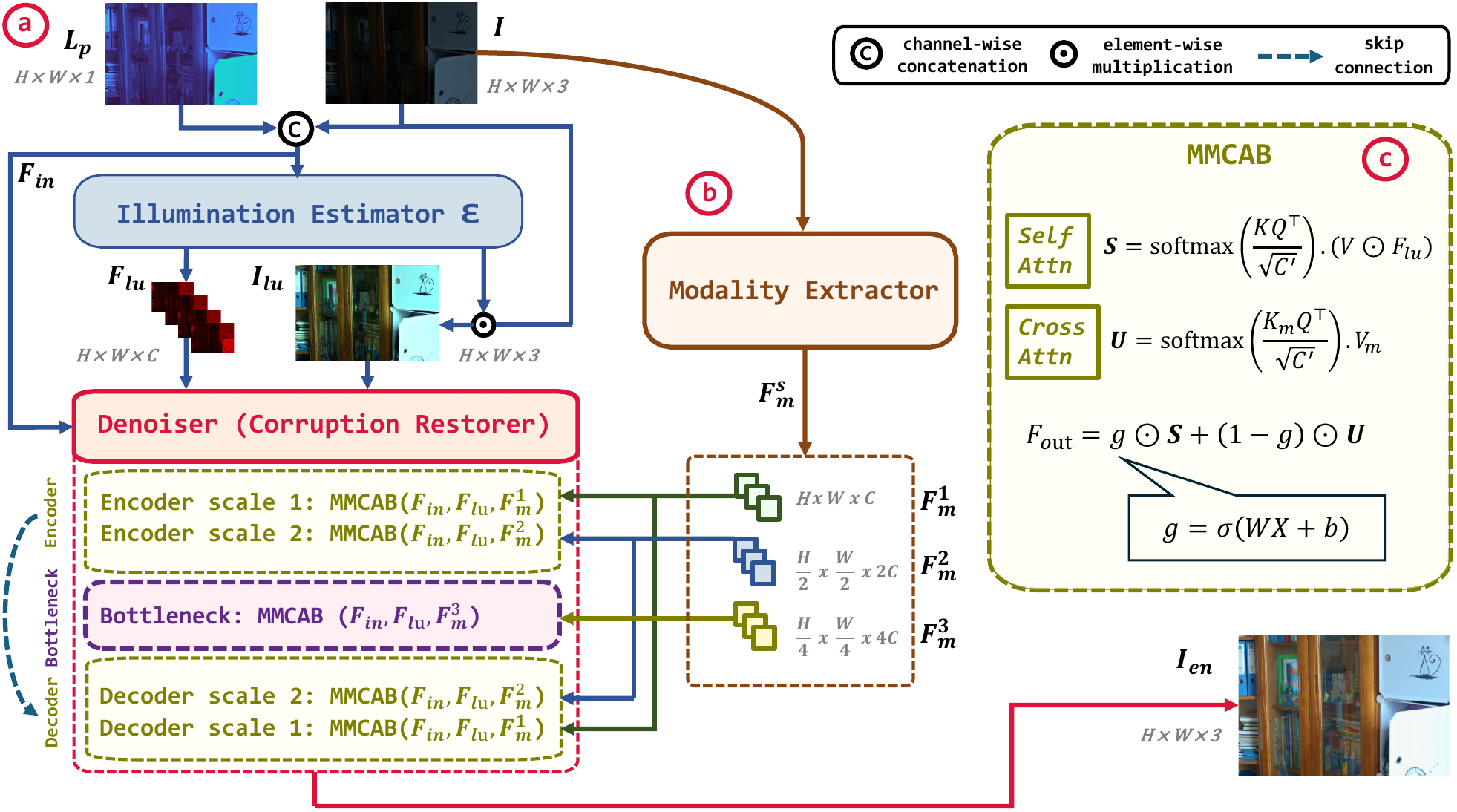}
	\vspace{-2mm}
	\caption{\small
	The overview of our M2Retinexformer architecture.
	(a) Illumination Estimator and Multi-Modal Corruption Restorer.
	(b) Modality Extractor that injects additional features into the corruption restorer.
	(c) The Multi-Modal Cross-Attention Block (MMCAB) that fuses RGB features $F_{in}$ with multi-modal features $F_{m}$ via cross-attention at different scales $s$.
	}
	\label{fig:architecture}
\end{figure*}
 
\noindent\textbf{Multi-Modal Learning}: Multi-Modal learning leverages complementary information from multiple modalities and has shown effectiveness across vision tasks. 
Depth estimation has been explored as an auxiliary modality for low-light image enhancement, demonstrating its effectiveness in modeling scene structure and illumination variation~\cite{wang2024multimodal}. Additionally, other approaches incorporate sensing modalities such as infrared or thermal imagery to improve illumination estimation~\cite{liu2025multi,wang2025thermal}.
ModalFormer~\cite{modalformer} proposed a multi-modal transformer for low-light enhancement that fuses diverse visual cues by leveraging the pre-trained 4M-21 model~\cite{bachmann20244m} to extract eight auxiliary modalities, but computational efficiency was not a primary design consideration.

Inspired by ModalFormer~\cite{modalformer}, our framework addresses the challenge of \textbf{enhancing Retinexformer by integrating only the most effective auxiliary modalities with minimal overhead}. We propose a hybrid architecture that builds upon Retinexformer’s illumination-guided restoration pipeline, while selectively incorporates auxiliary inputs using multi-modal cross-attention and adaptive gating mechanisms.

\section{METHOD}
\label{sec:method}
\vspace{-2mm}
As shown in Fig.~\ref{fig:architecture}, we present the overall architecture of M2Retinexformer, which extends Retinexformer by incorporating complementary multi-modal cues. The proposed framework introduces two main components:
Modality Extractor and Multi-Modal Cross-Attention Block (MMCAB).

% \begin{itemize}
%     \item Modality Extractor.
%     \item Multi-Modal Cross-Attention Block (MMCAB).
% \end{itemize}

\subsection{Preliminary: One-stage Retinexformer Framework}
\label{ssec:orf}
We adopt Retinexformer's one-stage Retinex-based framework (ORF) composed of an illumination estimator $\mathcal{E}$ and a corruption restorer $\mathcal{R}$. \\ 
Given a low-light image $\mathbf{I} \in \mathbb{R}^{H\times W\times 3}$ and its illumination prior map $\mathbf{L}_p \in \mathbb{R}^{H\times W}$ ($F_{in} = \left[I,\, L_p\right]$) :
\begin{equation}
(\mathbf{I}_{lu}, \mathbf{F}_{lu}) =  \mathcal{E}(\mathbf{I}, \mathbf{L}_{p}), ~~~~\mathbf{I}_{en} = \mathcal{R}(\mathbf{I}_{lu}, \mathbf{F}_{lu}),
\label{eq:orf}
\end{equation}
$\mathcal{E}$ takes $\mathbf{I}$ and $\mathbf{L}_p$ as inputs, then outputs the lit-up image $\mathbf{I}_{lu} \in \mathbb{R}^{H\times W\times 3}$ and lit-up features $\mathbf{F}_{lu} \in \mathbb{R}^{H\times W\times C}$, after that, $\mathbf{I}_{lu}$, $\mathbf{F}_{in}$ and $\mathbf{F}_{lu}$ are fed into $\mathcal{R}$ to suppress corruptions and produce the enhanced image $\mathbf{I}_{en} \in \mathbb{R}^{H\times W\times 3}$.

\subsection{Network Architecture}

\noindent\textbf{Illumination Estimator.} We retain Retinexformer's estimator, producing ${I}_{lu}$ and ${F}_{lu}$.

\noindent\textbf{Modality Extractor.} Modality features $F_m$ are extracted, aligned, and injected at multiple scales for cross-attention fusion with RGB features $F_{in}$.

\noindent\textbf{Multi-Modal Corruption Restorer.} The restorer follows a U-shaped encoder-decoder architecture. The proposed MMCAB augments Retinexformer's illumination-guided self-attention with multi-modal cross-attention.

\noindent\textbf{Adaptive Gating.} Gating balances illumination-guided self-attention from 
the RGB input and cross-attention from the auxiliary modalities based on modality reliability. 

\noindent\textbf{Progressive Refinement.} We cascade $\tau \in \{1,2,3\}$ identical refinement stages. Modality features are extracted once and reused across stages to reduce computational overhead.

\subsection{Modality Extractor}
\vspace{-2mm}
To overcome the limitations of RGB-only enhancement, we incorporate complementary auxiliary modalities such as:

\noindent\textbf{(i) Depth.} 
Depth provides illumination-invariant geometric structure that helps disambiguate dark regions caused by shadows, occlusions, or distance. We employ a frozen Depth-Anything-V2~\cite{yang2024depth} model to extract intermediate ViT features that serve as geometric priors.

\noindent\textbf{(ii) Luminance.}
Augmented luminance uses NTSC luminance, $L = 0.299 I_R + 0.587 I_G + 0.114 I_B$, where $I_R$, $I_G$, and $I_B$ are the RGB channels enriched with Sobel edges, local contrast, and multi-scale pyramid cues from the same input.

\noindent\textbf{(iii) Semantic Features.} 
To provide high-level contextual guidance, we extract semantic features using a frozen DINOv3~\cite{simeoni2025dinov3} backbone, which captures object-aware representations that help preserve color consistency and structural integrity in semantically complex regions.

For each modality $m$, features are extracted at multiple scales $s \in \{0,1,2\}$ and projected into a unified feature representation $F_m^s \in \mathbb{R}^{\frac{H}{2^s} \times \frac{W}{2^s} \times 2^sC}$ aligned with $F_{in}$.
The modality extractor follows a modular and extensible design, where each modality adheres to a unified interface. Adding a new modality requires registering it and implementing a lightweight encoder for that modality that conforms to the defined modality-extractor interface, keeping the framework extensible without modifying the core network.

\begin{table*}[htb]
	\centering
	\setlength\tabcolsep{6pt}
	\resizebox{0.995\textwidth}{!}{\hspace{-0.5mm}
		\begin{tabular}{l|cc|cc|cc|cc|cc|cc|cc}
			\toprule[0.15em]
			\multirow{2}{*}{Methods} 
			& \multicolumn{2}{c|}{LOL-v1} 
			& \multicolumn{2}{c|}{LOL-v2 real} 
			& \multicolumn{2}{c|}{LOL-v2 syn} 
			& \multicolumn{2}{c|}{SID} 
			& \multicolumn{2}{c|}{SMID} 
			& \multicolumn{2}{c|}{SDSD-in} 
			& \multicolumn{2}{c}{SDSD-out} \\
			& PSNR & SSIM & PSNR & SSIM & PSNR & SSIM & PSNR & SSIM & PSNR & SSIM & PSNR & SSIM & PSNR & SSIM \\ 
			\midrule[0.15em]
			RetinexNet~\cite{retinex_net} 
			& 16.77 & 0.560 & 15.47 & 0.567 & 17.13 & 0.798 & 16.48 & 0.578 & 22.83 & 0.684 & 20.84 & 0.617 & 20.96 & 0.629 \\

			KinD~\cite{kind} 
			& 20.86 & 0.790 & 14.74 & 0.641 & 13.29 & 0.578 & 18.02 & 0.583 & 22.18 & 0.634 & 21.95 & 0.672 & 21.97 & 0.654 \\

			Restormer~\cite{restormer} 
			& 22.43 & 0.823 & 19.94 & 0.827 & 21.41 & 0.830 & 22.27 & 0.649 & 26.97 & 0.758 & 25.67 & 0.827 & 24.79 & 0.802 \\

			MIRNet~\cite{mirnet} 
			& 24.14 & 0.830 & 20.02 & 0.820 & 21.94 & 0.876 & 20.84 & 0.605 & 25.66 & 0.762 & 24.38 & 0.864 & 27.13 & 0.837 \\

			Retinexformer~\cite{retinexformer} 
			& 23.61 & 0.836 & 21.85 & 0.839 & 25.52 & 0.929 
			& \textcolor{blue}{24.44} & \textcolor{red}{0.680} 
			& \textcolor{red}{29.01} & \textcolor{red}{0.814} 
			& 28.74 & 0.883 
			& \textcolor{blue}{27.95} & \textcolor{blue}{0.863} \\

			RetinexMamba~\cite{RetinexMamba} 
			& 24.03 & 0.827 
			& \textcolor{blue}{22.45} & 0.844 
			& \textcolor{blue}{25.89} & \textcolor{blue}{0.935} 
			& -- & -- & -- & -- & -- & -- & -- & -- \\

            SNR-Net~\cite{snr_net} 
			& \textcolor{blue}{24.61} & \textcolor{blue}{0.842} 
			& 21.48 & \textcolor{blue}{0.849} 
			& 24.14 & 0.928 
			& 22.87 & 0.625 
			& 28.49 & \textcolor{blue}{0.805}
			& \textcolor{blue}{29.44} & \textcolor{blue}{0.894} 
			& \textcolor{red}{28.66} & \textcolor{red}{0.866} \\

			\midrule[0.15em]
			\textbf{M2Retinexformer} 
			& \textcolor{red}{24.89} & \textcolor{red}{0.859} 
			& \textcolor{red}{23.85} & \textcolor{red}{0.891} 
			& \textcolor{red}{27.12} & \textcolor{red}{0.950} 
			& \textcolor{red}{24.84} & \textcolor{blue}{0.678} 
			& \textcolor{blue}{28.76} & \textcolor{red}{0.814} 
			& \textcolor{red}{30.48} & \textcolor{red}{0.908} 
			& 27.39 & 0.843 \\
			\bottomrule[0.15em]
	\end{tabular}}
	\caption{Quantitative comparisons on LOL v1/v2, SID, SMID, and SDSD datasets. Best results in \textcolor{red}{red} and second-best in \textcolor{blue}{blue}.}
	\label{tab:quantitative}
\end{table*}

\vspace{-2mm}
\subsection{Multi-Modal Cross-Attention Block (MMCAB)}
The MMCAB is the core fusion module that integrates RGB features with auxiliary modalities via cross-attention.

\noindent\textbf{Multi-Modal Cross-Attention.} 
Given RGB features $F_{in}$ and modality features $F_m^s$ at scale $s$, we reshape them into tokens $X, X_m \in \mathbb{R}^{N \times C'}$ with $N = H'W'$.
Queries are derived from RGB features, while keys and values are obtained from the auxiliary modality:

\vspace{-5mm}
\begin{equation}
Q = XW_Q, \quad K_m = X_m W_{K_m}, \quad V_m = X_m W_{V_m},
\end{equation}
with $Q, K_m, V_m \in \mathbb{R}^{N \times C'}$, and $W_Q, W_{K_m}, W_{V_m} \in \mathbb{R}^{C' \times C'}$ are learnable projection matrices. The resulting cross-attention  ${\text{A}_m} \in \mathbb{R}^{N \times C'}$ for modality $m$ is computed as:

\vspace{-1mm}
\begin{equation}
\text{A}_m =
\text{softmax}\left(\frac{Q K_m^\top}{\sqrt{C'}}\right) V_m,
\end{equation} 
allowing RGB features to selectively query complementary information from auxiliary modalities.

\noindent\textbf{Illumination-Guided Self-Attention.} 
In parallel, self-attention is applied to RGB features, where queries, keys, and values are all derived from the same source $X$:

\vspace{-2mm}
\begin{equation}
Q = XW_Q, \quad K = XW_K, \quad V = XW_V,
\end{equation}
with $Q, K, V \in \mathbb{R}^{N \times C'}$.
Following Retinexformer, the value features are modulated by illumination cues $F_{lu} \in \mathbb{R}^{N \times C'}$:

\vspace{-3mm}
\begin{equation}
A = \text{softmax}\left(\frac{QK^\top}{\sqrt{C'}}\right), \quad
S = A \left(V \odot F_{lu}\right),
\end{equation}
where $A \in \mathbb{R}^{N \times N}$ denotes the attention weight matrix and
$S \in \mathbb{R}^{N \times C'}$ is the resulting illumination-guided self-attention output.
This design encourages the attention mechanism to focus on relevant features in the RGB input.

\begin{figure*}[htb]
	\centering
	\includegraphics[
    width=0.9\textwidth
	]{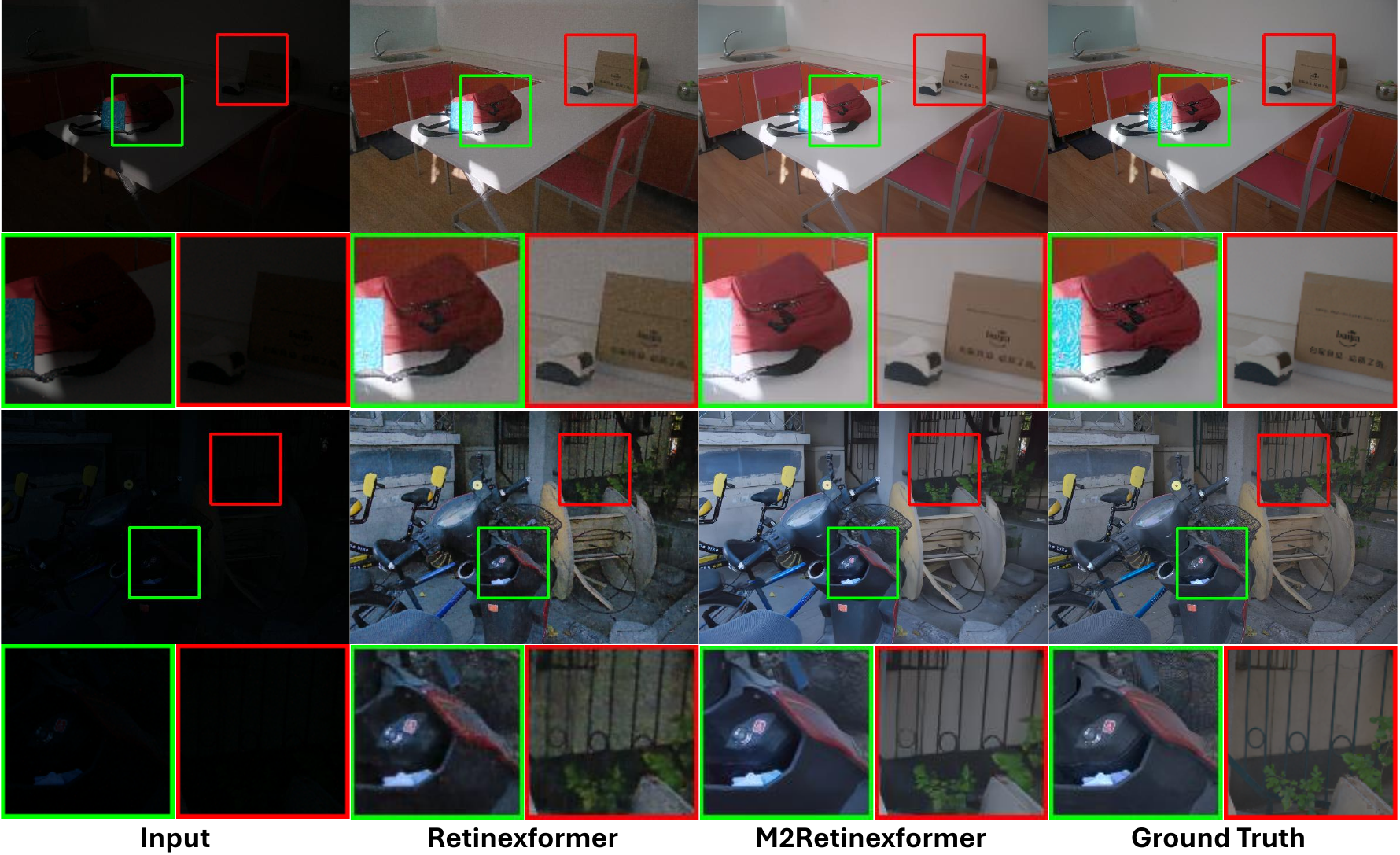}

    \vspace{-2mm}
	\caption{\small
	Visual results on LOL-v2 Real. Our M2Retinexformer enhances visibility while preserving color fidelity and suppressing noise.
	}
    \vspace{-2mm}
	\label{fig:qualitative}
\end{figure*}

\noindent\textbf{Adaptive Gating.} Cross-attention output for each modality ${\text{A}_m}$ is weighted by a learnable gate $g_m$ based on its reliability:

\vspace{-3mm}
\begin{equation}
U = \sum_{m} g_m \odot \text{A}_m, \quad g_m = \sigma(W_m X + b_m),
\end{equation}
\vspace{-3mm}

This multi-modal output $U \in \mathbb{R}^{N \times C'}$ is then combined with the self-attention output $S$ via a final gate $g_f$ that balances illumination-guided self-attention with multi-modal cross-attention:
\vspace{-2mm}
\begin{equation}
\text{Output} = g_f \odot S + (1-g_f) \odot U, \quad g_f = \sigma(W_f X + b_f),
\end{equation} 
where $W_m$, $W_f$, $b_m$ and $b_f$ are learnable.

\noindent\textbf{MMCAB Structure.}
\begin{equation}
\begin{aligned}
F' &= F_{in} + \text{MMCAB}(\text{LN}(F_{in}), F_{lu}, \{F_m\}), \\
F_{out} &= F' + \text{FFN}(\text{LN}(F')),
\end{aligned}
\end{equation}
The block follows a residual design. LN denotes layer normalization and FFN is a feed-forward network. In the final stage $F_{out} \in \mathbb{R}^{H' \times W' \times C'}$ is projected to the RGB space, producing $I_{\text{en}} \in \mathbb{R}^{H \times W \times 3}$.

\vspace{-3mm}
\subsection{Loss Function}

Retinexformer originally employed only the L1 loss. We find that incorporating a \textbf{perceptual loss}~\cite{johnson2016perceptual} improves visual quality and preserves high-level semantic structures and textures that are relevant for low-light enhancement, where fine details can easily be lost during brightness adjustment.
The combined objective is:
\vspace{-2mm}
\begin{equation}
\mathcal{L} = \mathcal{L}_{1} + \lambda_{per}\mathcal{L}_{per},
\end{equation}
where $\mathcal{L}_{1} = \|\mathbf{I}_{en}-\mathbf{I}_{gt}\|_{1}$ is the L1 loss between $\mathbf{I}_{en}$ and the ground truth image. $\mathcal{L}_{per}$ is a VGG-19 perceptual loss. We set $\lambda_{per}=0.5$ based on validation performance.

\vspace{-2mm}
\section{EXPERIMENTS}
\label{sec:experiments}
\vspace{-2mm}
\subsection{Experimental Setup and Implementation Details}

\noindent\textbf{Datasets.} We evaluated M2Retinexformer on seven low-light benchmarks: LOL-v1~\cite{retinex_net}, LOL-v2 Real/Synthetic~\cite{lol_v2}, SID~\cite{sid}, SMID~\cite{smid}, and SDSD Indoor/Outdoor~\cite{sdsd}.

\noindent\textbf{Training details.} Our framework is implemented in PyTorch and trained using the Adam optimizer. For each dataset, training is performed until convergence with a dynamically adjusted learning rate using either Cosine Annealing or Reduce-on-Plateau scheduling. Batch and patch sizes are selected separately for each dataset, and standard data augmentation is applied. Performance is evaluated using PSNR and SSIM. The complete configs, train/eval scripts, and checkpoints are released alongside the code to ensure reproducibility.

\noindent\textbf{Model complexity.} 
M2Retinexformer has 2M trainable params and 48M total params, including frozen Depth-Anything-V2 and DINOv3 encoders that do not add optimization complexity. This is about \textbf{1/4} of the 4M-21 extractor~\cite{bachmann20244m} (198M params) used in ModalFormer~\cite{modalformer}. All experiments are conducted on a single NVIDIA RTX 5090 GPU.

\vspace{-3mm}
\subsection{Performance Evaluation}

\noindent\textbf{Quantitative results.} Table~\ref{tab:quantitative} compares our method with several recent approaches. M2Retinexformer achieves the best or second-best performance on most benchmarks, demonstrating the robustness and applicability of the proposed architecture, as well as the effectiveness of the multi-modal fusion and reliability-aware gating strategy that balances self-attention and cross-attention. The lower PSNR gains on SMID and SDSD are likely due to their video-based short/long-exposure captures, which exhibit different exposure characteristics and degradation patterns, making auxiliary modalities less stable. ModalFormer is the closest related work; however, we do not include it in Table~\ref{tab:quantitative} due to the lack of publicly available code and reproducible results. All experiments are conducted without GT Mean correction for fair comparison.

\noindent\textbf{Qualitative Results.} Visual comparisons in Fig.~\ref{fig:qualitative} show that Retinexformer suffer from color distortion or residual noise, whereas M2Retinexformer produces well-exposed images with natural colors and reduced noise, benefiting from the injected modalities and perceptual loss.

\subsection{Ablation Study}

We conducted a comprehensive ablation study on the LOL-v2 Real dataset to quantify each component’s contribution and validate our design choices. As shown in Table~2, under perceptual loss supervision, depth yields the most significant performance gains, followed by luminance. The results also show that adding all modalities does not consistently improve performance, demonstrating that effective modality selection remains critical in multi-modal enhancement. Although adaptive gating is designed to do its best to suppress unnecessary modalities, it cannot fully offset the interaction between noisy or redundant cues and the main RGB branch.

\begin{table}[htb]
	\centering
    \scriptsize
	\resizebox{0.48\textwidth}{!}{
		\begin{tabular}{c c c c c | c c}
			\toprule
			Baseline & $Perc_{loss}$ & DinoV3 & Depth & Luminance & PSNR & $\Delta$PSNR \\
			\midrule
			\checkmark &  &  &  &  & 21.85 & -- \\
			\checkmark & \checkmark &  &  &  & 22.81 & +0.96 \\
            \checkmark & \checkmark & \checkmark &  &  & 22.90 & +1.05 \\
			\checkmark & \checkmark &  & \checkmark &  & \textbf{23.85} & \textbf{+2.00} \\
            \checkmark & \checkmark &  &  & \checkmark & 23.29 & +1.44 \\
            \checkmark & \checkmark & \checkmark & \checkmark &  & 22.93 & +1.08 \\
            \checkmark & \checkmark & \checkmark &  & \checkmark & 23.67 & +1.82 \\
            \checkmark & \checkmark &  & \checkmark & \checkmark & 22.35 & +0.50 \\
            \checkmark & \checkmark & \checkmark & \checkmark & \checkmark & 23.35 & +1.50 \\
			\bottomrule
		\end{tabular}
	}
    \vspace{-3mm}
	\caption{\small Ablation study on LOL-v2 Real using $\tau$=3 showing the contribution of each component. Gain indicates the absolute PSNR improvement over the baseline without any additional components.}
	\label{tab:ablation_components}
\end{table}

\vspace{-4mm}
\section{CONCLUSION}
\label{sec:conclusion}
\vspace{-2mm}
In this paper, we propose M2Retinexformer, a multi-modal extension of Retinexformer that incorporates heterogeneous modalities through cross-attention fusion. Our key insight is that depth provides geometric context that is robust to illumination changes, while luminance and semantic features provide content-aware guidance. Integrated through the proposed MMCAB, these modalities improve enhancement quality. Evaluations across multiple benchmarks show that our model provides overall performance gains over existing methods. A limitation is that the benefits of multi-modal fusion depend on modality reliability, and gains may diminish when auxiliary features are unstable.
The proposed framework further provides a modular and extensible design that can accommodate additional priors, making it a promising direction for future advances in low-light image enhancement.

\bibliographystyle{IEEEbib}
\small
\bibliography{references}

\end{document}